# Research on the Application of Deep Learning-based BERT Model in Sentiment Analysis

Yichao Wu[1*], Zhengyu Jin[1], Chenxi Shi[2], Penghao Liang[3], Tong Zhan[4]

1*Software development,Northeastern University,Boston, MA,USA
1 Informatics,Univeristy of California, Irvine,Irvine ,CA,USA
2 Software development ,Telecommunication Systems Management ,Northeastern University,Boston, MA,USA
3Information Systems, Northeastern University,Boston, MA,USA
4 Computer Science,Columbia University,NY, USA
*Corresponding author:Yichao Wu[E-mail:wu.yicha@northeastern.edu]

## ABSTRACT

This paper explores the application of deep learning techniques, particularly focusing on BERT models, in sentiment analysis. It begins by introducing the fundamental concept of sentiment analysis and how deep learning methods are utilized in this domain. Subsequently, it delves into the architecture and characteristics of BERT models. Through detailed explanation, it elucidates the application effects and optimization strategies of BERT models in sentiment analysis, supported by experimental validation. The experimental findings indicate that BERT models exhibit robust performance in sentiment analysis tasks, with notable enhancements post fine-tuning. Lastly, the paper concludes by summarizing the potential applications of BERT models in sentiment analysis and suggests directions for future research and practical implementations.

## 1. INTRODUCTION

With the rapid development of the Internet and social media, a large amount of text data has been produced, which contains rich emotional information. Sentiment analysis, as a technique designed to identify and process people's emotional tendencies, emotions and perspectives, has become an important research direction in the field of natural language processing (NLP). It plays an important role in many fields such as product review analysis, public opinion monitoring, and consumer behavior prediction, helping companies and organizations better understand public sentiment and make more informed decisions.

In the research and application of sentiment analysis, the rise of deep learning technology provides new methods and perspectives for processing complex text data. Compared to traditional rule-based or machine learning methods, deep learning can automatically learn features on large-scale data sets, significantly improving the accuracy and efficiency of sentiment classification and analysis. Among them, BERT (Bidirectional Encoder Representations from Transformers) model, as an advanced deep learning framework, captures deep semantic information in text through pre-training. Breakthrough results have been achieved in multiple NLP missions.

The BERT model has attracted a lot of attention in the field of natural language processing, mainly because of its bidirectional Transformer architecture, which can more fully understand the complex relationships in the language context. This comprehensive understanding enables BERT to excel in tasks such as sentiment analysis, where he is able to more accurately capture the emotional tendencies of text. However, while the BERT model has been proven to be effective in a number of fields, its effectiveness and best practices for specific sentiment analysis tasks remain to be explored.

In view of this, the purpose of this study is to explore the application effect and method of BERT model in sentiment analysis. Through in-depth analysis and experimental verification of the BERT model, we expect to reveal its effectiveness and potential limitations in identifying different text emotional tendencies, providing guidance for future research and practice. This paper will first review the basic concepts of sentiment analysis and the application progress of deep learning in this field, then introduce the architecture and characteristics of BERT model in detail, and finally verify the application effect and optimization strategy of BERT model in sentiment analysis through experimental research.

## 2. RELATED WORK

### 2.1 Sentiment analysis review

Sentiment Analysis is the process of revealing people's opinions, positions, or attitudes toward a topic, person, or entity. In the early studies, the object of sentiment analysis was often only the text content, and text analysis alone has certain disadvantages, that is, it only uses words and phrases that are not enough to extract accurate opinions as clues, which may lead to problems such as differences in language meaning, language sparsity, and insufficient connection with real context.

With the development of social media and the increasing richness of information content and carriers, the construction of multimodal information such as text, vision and hearing provides a new opportunity for the development of sentiment analysis. Nowadays, more and more people like to record their feelings about the use of a product, the evaluation of a brand, or the opinion of a social event through video or audio, and share it on social media. Compared with text, video has great potential for diversity in content creation, and is often more attractive and convincing, so its influence and radiation are also wider. In such video or audio, the facial expressions of the characters, the voice intonation of the human voice, the melody of the background music, and even physiological signals such as brain electricity are all effective tools to convey emotional information.

Therefore, in recent years, multimodal sentiment analysis has become a hot research direction in the field of natural language processing and computer vision. It can be predicted that in the future for a long time, this multimodal sentiment analysis method will become the main means of multimedia content analysis and understanding, and can be widely used in a variety of scenarios, such as stock market prediction, election polls, customer satisfaction assessment, news public opinion perception, advertising recommendation, etc.

### 2.2 Application of deep learning to NLP

Deep learning has become the dominant method of natural language processing (NLP) research, especially in large-scale corpora. In natural language processing tasks, sentences are often thought of as a series of markers. Therefore, popular deep learning techniques such as recurrent neural networks (RNN) and convolutional neural networks (CNN) have been widely used in text sequence modeling.

However, there are a large number of natural language processing problems that can best be expressed using graph structures. For example, structural and semantic information in sequence data (e.g., various syntactic analysis trees (e.g., dependency analysis trees) and semantic analysis graphs (e.g., abstract meaning representation graphs) can augment the original sequence data by incorporating task-specific knowledge. As a result, these graphically structured data can encode complex paired relationships between entity tags to learn more information representations. However, deep learning techniques are known to be disruptive to Euclidean data (such as images) or sequential data (such as text), but are not immediately applicable to graph-structured data. As a result, this gap has driven research into deep learning for graphs, particularly the development of graph neural networks (GNN).

This wave of research at the intersection of graph deep learning and natural language processing has influenced a variety of natural language processing tasks. There has been a surge of interest in applying/developing various types of GNN and considerable success in many natural language processing tasks, from classification tasks such as sentence classification, semantic role labeling, and relationship extraction, to generation tasks such as machine translation, question generation, and summarization.

## 2.3 Bidirectional Encoder Representations from Transformer

The full name of BERT model is: Bidirectional Encoder Representations from Transformer. As can be seen from the name, the goal of the BERT model is to use large-scale unlabeled corpus to train and obtain a Representation of a text containing rich semantic information, that is, a semantic representation of the text, and then fine-tune the semantic representation of the text in a specific NLP task, and finally apply it to that NLP task. Among them, the Attention mechanism is also the key part of the analysis. In essence, Attention is to select a small amount of important information from a large amount of information, and focus on these important information, ignoring most of the unimportant information. The greater the weight, the more it focuses on its corresponding Value, that is, the weight represents the importance of information, and Value is its corresponding information.

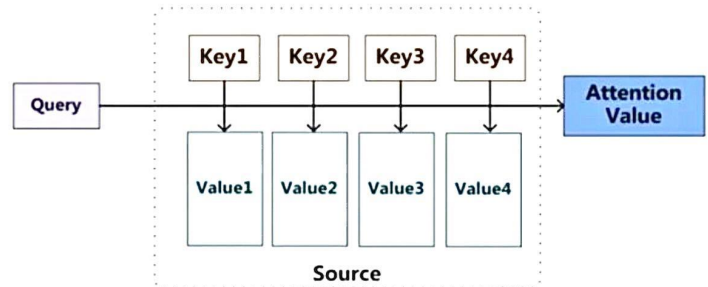

Figure 1. BERT model Attention mechanism architecture diagram

As for the specific calculation process of the Attention mechanism, if most of the current methods are abstracted, it can be summarized into two processes: the first process is to calculate the weight coefficient according to Query and Key, and the second process is to perform weighted summation of Value according to the weight coefficient. The first process can be subdivided into two stages: the first stage calculates the similarity or correlation between Query and Key; In the second stage, the original score of the first stage is normalized. In this way, the Attention calculation process can be abstracted into the three stages shown in the figure.

# 3. BERT MODEL FOUNDATION AND METHODOLOGY

## 3.1 Model architecture

BERT (Bidirectional Encoder Representations from Transformers) is a pre-trained natural language processing model released by Google in 2018. At the heart of the BERT model is the Transformer encoder, which allows unsupervised pre-training on a large corpus and then fine-tuning on a variety of NLP tasks through fine-tuning. BERT model is a bidirectional deep learning model that considers all words in context at the same time to better understand the meaning of a sentence. BERT models have been shown to achieve the best results on a number of NLP tasks, including question answering, text classification, and named entity recognition.

- BERT is a natural language understanding model based on deep neural networks that can learn the semantics and structure of a language from large-scale unlabeled text.
- BERT's innovation lies in its use of a bi-directional Transformer encoder that can take into account context information in both left and right directions to capture richer language features.
- BERT's pre-training tasks include MLM and NSP, which are used to learn vocabulary and sentence-level representations, respectively. MLM is a cloze task that randomly replaces some words in the input text with a special symbol [MASK] and then asks the model to predict the obscured words. NSP is a binary classification task that gives two sentences and lets the model decide whether they are continuous or not.
- BERT has achieved significant improvements in a variety of natural language processing tasks, such as question answering, sentiment analysis, named entity recognition, and text classification. BERT not only improves the performance of the model, but also simplifies the fine-tuning process of the model, which can be adapted to different tasks by adding only a small number of task-related layers on top of the pre-trained model.
- BERT has also spawned many improved or extended models based on it, such as RoBERTa, ALBERT, XLNet, ELECTRA and others. These models optimize or innovate BERT in different ways, such as increasing the amount of data, reducing the number of parameters, and changing the pre-training task.

## 3.2 Pre-training task

In BERT, language models are built in two pre-trained ways:

1) BERT language model Task 1: MASKED LM

In BERT, Masked LM(Masked language Model) built a language model, which was also one of BERT's pre-training tasks. Simply put, masked or replaced any word or word in a sentence randomly. Then, the model is asked to predict the covered or replaced part through understanding the context, and then only calculate the Loss of the covered part when making Loss, which is actually an easy task to understand, and the actual operation is as follows:

Randomly replace 15 tokens in a sentence with the following:

1. These tokens have an 80 chance of being replaced with mask;

2. There is a 10% chance that it will be replaced with any other token;

3. There's a 10% chance it's intact.

Then let the model predict and restore the hidden or replaced part, the final output of the hidden layer of the model's calculation results are:

$$X_{hidden} : batch\_size, seq\ len, embedding\_dim \quad (1)$$

This step requires initializing a mapping layer weight $W_{vocab}$:

$$W_{vocab} : [embedding\_dim, vocab\_size] \quad (2)$$

In use Wocab to complete the mapping of hidden dimensions to the number of word vectors, requiring only the matrix multiplication (dot product) of $X_{xdden}$ and $W_{vocab}$ :

$$X_{hidden}W_{vocab}: [batchsize, segien] \quad (3)$$

After vocabsize, the above calculation results are normalized in theVocab_size(the last) dimension so.ftmax is the sum of vocab_size corresponding to each word to 1. Then we can obtain the prediction results of the model through the word with the highest probability in $V_{cocab}$_size, and we can compare it with the prepared La bel makes a Loss and backpasses the gradient.

2)BERT Language Model Task 2 :Next Sentence Prediction

1. First we get a pair of sentences that belong to the context, that is, two sentences, and then we add some special tokens to these two consecutive sentences:[cls]

In the beginning of the sentence, and in the end of the sentence, as in figure 2:

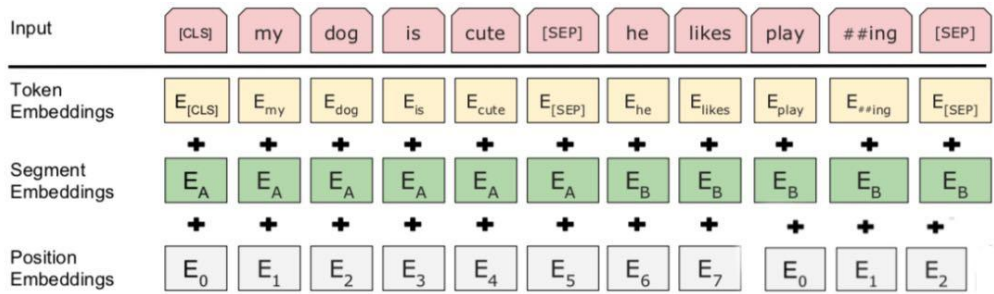

Figure 2. BERT sentiment analysis model

2. After completing the above steps, a trainable segment emmbeddings should be randomly initialized. As shown in the figure above, the function is to separate the next sentence with the information of embeddings. Let the model determine where the next sentence starts and ends, for example:

[cls]0 and 1 above are segment embeddings.

### 3.3 Fine-tuning mechanism

1. Pre-processing of sentiment analysis corpus: Refer to the directory./corpus/sentiment_preprocessing.ipynb. The corpus size of the hotel review corpus used in this study is larger than that of the sentiment analysis conducted by LSTM in 2018, with 5000 positive comments and 5000 negative comments respectively. In fact, this is also a toy data set, with such a large BERT parameter model, training will produce serious overfitting and poor generalization ability, which is also the following problem we need to solve:

2. In the Next Sentence Prediction of BERT's training, we took the vector corresponding to [cls], mapped it to a value and activated it with sigmoid function:

$$\hat{y} = sigmoid(Linear(cls\_vector)) \quad \hat{y} \in (0, 1) \quad (4)$$

g=sigmoid(Linear(cls_vector))gE(0, 1)3. Dynamic learning rate and early stop:

In the previous step we divided the corpus into training and test sets, and we trained each epoch with the training set. The measure of model performance is the AUC, which is very easy to use for binary classification.

After training the current epoch, measure the current training results against the test set, and record the AUC of the current epoch. If the current AUC does not improve from the previous epoch, reduce the learning rate. In practice, reduce the current learning rate by 1/5, and terminate the training until the AUC of the 10 epoch test sets does not improve

Our initial learning rate is 1e-6, because we are training on the basis of the Wikipedia pre-training corpus, which is a downstream task and only needs to fine-tune the pre-training model.

5. Threshold fine-tuning: After inference of the model, the output value is between 0 and 1, so in the process of implementing BERT model fine-tuning, it can be considered that as long as the value is above 0.5, it is a positive sample, and if it is below 0.5, it is a sub-sample. In fact, this is not necessarily,0.5 is usually not the best classification boundary.

The method of this script is to define 99 thresholds from 0.01 to 0.99. If the threshold is higher than the positive sample and lower than the sub-sample, f1 score is calculated with the test set, and then the threshold that can make f1 score highest is selected. In the training, each epoch will run a script to find the threshold value, as shown in the following table:

Table 1. Fine-tune the threshold data table

| Epoch | Train Loss | Train AUC | Test Loss | Test AUC |
| --- | --- | --- | --- | --- |
| 414 | 0.274383 | 0.954070 | 0.283045 | 0.958663 |
| 415 | 0.280170 | 0.952098 | 0.283048 | 0.958663 |
| 416 | 0.279494 | 0.952490 | 0.283052 | 0.958663 |
| 417 | 0.277347 | 0.953116 | 0.283056 | 0.958663 |
| 418 | 0.278657 | 0.952766 | 0.283057 | 0.958663 |

- According to the tabular data, changes in different rounds (Epochs), training Loss (Train Loss) and training AUC (Area Under Curve) as well as Test loss (Test Loss) and test AUC during the training process can be observed. AUC is an indicator to measure the performance of a classifier, and the closer the value is to 1, the better the performance.
- During the Fine-tune threshold process, the model performance is evaluated by adjusting different thresholds (from 0.01 to 0.99) and calculating an F1 score. The F1 score combines the accuracy and recall rate of the model to evaluate the overall performance of the model.
- It can be seen from the table data that with the increase of Epochs, training Loss and Test Loss remain relatively stable, while AUC values (Train AUC and Test AUC) gradually increase, indicating that the performance of the model on the training and test set continues to improve.
- Through the process of Fine-tune thresholds, the threshold that makes F1 score the highest can be selected to achieve the best model performance.

### 3.4 Adaptation and optimization of BERT model

In the process of realizing the adaptive optimization of BERT model, the following methods need to be adopted:

1) Improve the hidden language model

In BERT model, text preprocessing is segmented according to the smallest unit. For example, the pre-processing of English text uses Google's wordpiece method to solve the problem of unknown words.

The object covered in MLM is mostly a subword, not a full word; For Chinese, it is directly segmented according to the word, and directly covers the single word. This masking strategy leads to incomplete learning of word information in the model. In response to this shortcoming, most researchers have improved the MLM masking strategy. In the BERT-WWM model subsequently released by Google, the way of full word coverage is proposed.

BERT-Chinese-wwm uses Chinese word segmentation to cover all the words that make up a complete word at the same time.

ERNIE extends the Chinese full-word masking strategy to cover Chinese word segmentation, phrases and named entities.

SpanBERT adopted geometric distribution to randomly sample the masked phrase fragments, and used Span boundary word vectors to predict the masked words

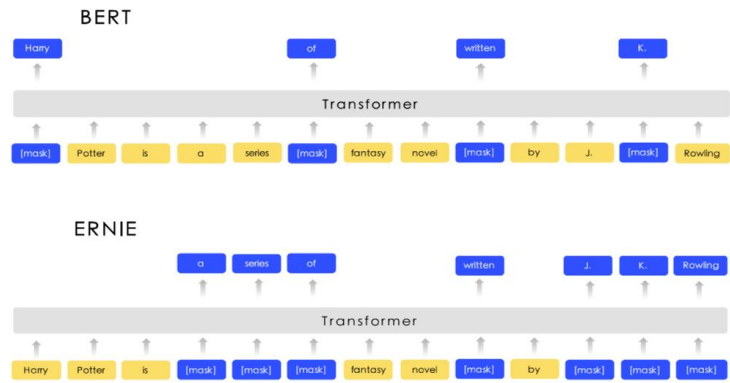

Figure 3. The different masking strategy between BERT and ERNIE

2) Noise reduction autoencoder is introduced

MLM randomly replaces the words in the original text with [MASK] tags, which itself destroys the text, equivalent to adding noise to the text, and then trains the language model to restore the text and eliminate the noise.

DAE is an autoencoder with noise reduction function, which is designed to restore the input data containing noise to clean raw data. For the language model, it is to add the noise data to the original language, and then remove the noise through the model learning to restore the original text.

BERT introduces the noise reduction autoencoder, which enriches the ways of text destruction. For example, random masking (consistent with MLM) certain words, random deletion of certain words or fragments, shuffling the order of documents, etc., after the text is input into the encoder, a decoder is used to generate the original document before the destruction.

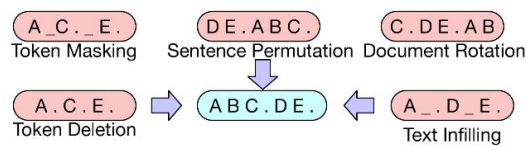

3) Introducing alternative word detection

MLM makes predictions about the words marked by [MASK] in the text in an attempt to recover the original text. Its prediction may be completely correct, or it may predict a word that is not part of the original text.

ELECTRA introduces alternative word detection to predict which words in a sentence generated by a language model are words in the original sentence and which words are words generated by the language model and do not belong to the original sentence.

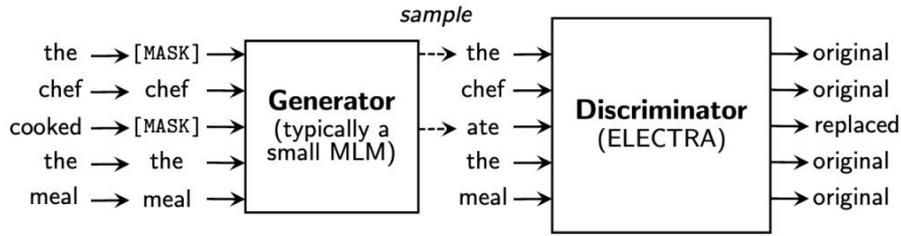

ELECTRA uses a small MLM model as a Generator to make predictions about sentences containing [MASK]. In addition, a Discriminator based on binary classification is trained to judge the sentences generated by the generator.

## 4. EXPERIMENT AND CONCLUSION

In the experiment, we will implement sentiment analysis methods using deep learning combined with BERT models and compare them with other traditional word vector models, including FastText, Word2Vec, and GloVe. We choose DistilBERT as a word embedding model for text sentiment classification, and evaluate the performance of DistilBERT by comparing the performance of different models in sentiment analysis tasks. In the experiment, we will use public sentiment classification data sets for training and testing, and use accuracy, accuracy, recall and other indicators to evaluate the model. In order to achieve the goal of the experiment, we will import each word vector model in turn (including FastText, Word2Vec, GloVe, and DistilBERT) and show how to load the pre-trained model and apply it to the sentiment analysis task.

### 4.1 Data set

The SST2 dataset (Stanford Sentiment Treebank) is a common dataset for sentiment analysis tasks, which mainly contains the text of movie reviews and corresponding sentiment tags. Each comment was labeled as a positive emotion (1) or a negative emotion (0) to indicate the polarity of emotion expressed in the comment. This dataset has the following characteristics:

Review texts: The SST2 dataset contains a large number of film review texts from different channels and platforms, covering a variety of language styles and thematic content. The diversity of comment text enables the model to learn a wider range of language expression and emotional experience during training.

- - Emotional Tags: Each comment is equipped with an emotional tag that indicates the emotional attitude expressed by the comment. These sentiment labels employ a binary classification of positive emotion (1) and negative emotion (0), making the sentiment analysis task simple and intuitive.
- - Data balance: SST2 data set has a certain balance on positive and negative samples, which means that the number of comments on positive and negative emotions is relatively balanced, which helps the model to better understand and distinguish different emotional polarities in the learning process.
- - Quality of annotation: Because of the high quality of annotation in the SST2 dataset, it has become a standard benchmark dataset for sentiment analysis tasks. High-quality labeling helps to ensure that the model receives accurate supervision signals during training and evaluation, thereby improving the model's performance and generalization ability.

Table 2: The sentiment analysis table uses the label 0/1 to represent positive and negative emotions

| sentence | label |
| --- | --- |
| a stirring , funny and finally transporting re imagining of beauty and the beast and 1930s horror films | 1 |
| apparently reassembled from the cutting room floor of any given daytime soap | 0 |
| they presume their audience won't sit still for a sociology lesson | 0 |
| this is a visually stunning rumination on love , memory , history and the war between art and commerce | 1 |
| jonathan parker 's bartleby should have been the be all end all of the modern office anomie films | 1 |

SST2 dataset is widely used in the training and evaluation of sentiment analysis tasks, and is favored by researchers because it contains rich text data and accurate sentiment labels. In this experiment, we will use the SST2 dataset as training and test data to evaluate the performance of the proposed deep learning model on the emotion classification task.

## 4.2 Model building

The emotion classification model of sentences consists of two parts:

DistilBERT processes the input sentence and passes on some of the information it extracts from the sentence to the next model. DistilBERT is a smaller version of the BERT model that was open-source by the HuggingFace team. It retains BERT's power while being smaller and faster than BERT.

A basic Logistic Regression model that will process the output of DistilBERT and categorize the sentences, output 0 or 1.

The data passed between the two models is a 768 dimensional vector.

Assuming the sentence length is n, then a sentence passing through BERT should yield n 768 dimensional vectors.

Model training and prediction

1) Training

Two models are used in this article, but only the Logistic Regression model needs to be trained during the implementation.

For the DistilBERT model, the parameters that were pre-trained by the model were used, and the model was not used to train and fine-tune the sentence classification task.

Use the Scikit Learn toolkit to do so. Divide the entire BERT output into train/test datasets.

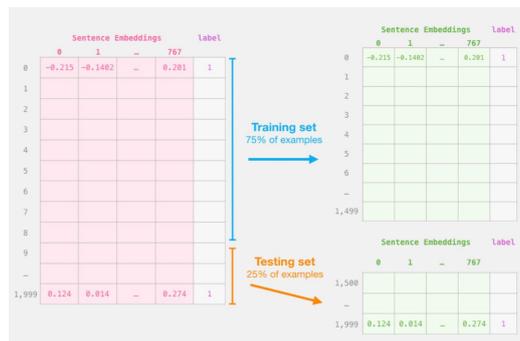

Figure 4. Step #2: Test/Train Split for model #2, logistic regression

- Divide 75% of the data into the training set and 25% into the test set.
- sklearn's train/test split shuffles the sample before splitting it.
- The next step is to train the regression model using machine learning methods.

## 4.3 forecast

How do you use models to make predictions?

For example, we want to categorize the sentence "a visually stunning rumination on love.

The first step is to use BERT's tokenizer to divide sentences into tokens;

Second, add special tokens for sentence classification tasks (add [CLS] at the beginning of the sentence and [SEP] at the end);

In the third step, the tokenizer will replace each token with the ID in the embedding table that comes with our pre-trained model.

In this experiment, we compared sentiment analysis methods using deep learning with BERT models against traditional word vector models like FastText, Word2Vec, and GloVe. We chose DistilBERT for text sentiment classification and evaluated its performance on the SST2 dataset. The model architecture included DistilBERT for processing input sentences and logistic regression for categorizing them. We split the data, trained the logistic regression model, and made predictions using tokenization and embedding mapping. Our goal is to assess DistilBERT's effectiveness compared to traditional models in sentiment analysis. Now, let's move on to the experimental results.

## 4.4 experimental results

BERT model training was carried out on the original data set. The output tensor size of BERT model was [2000, 59, 768]. # Vector corresponding to [CLS]token was extracted, and the results were as follows:

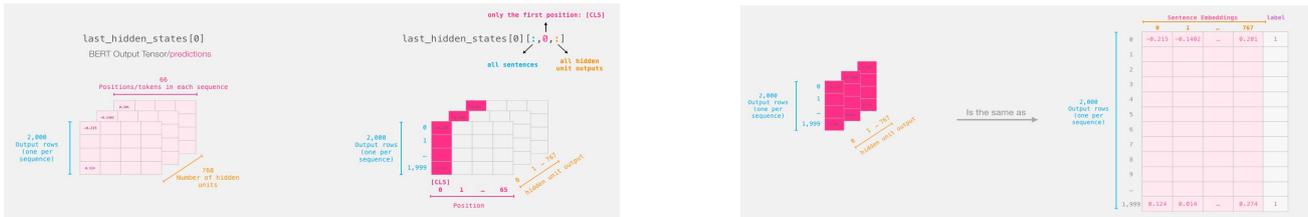

Figure 5. The tensor size of the BERT model output is [2000, 59, 768], the vector corresponding to the token

According to the above model, it can be found that the results of the trained classifier are obviously better than those of the random prediction.

But this example probably doesn't show BERT's performance at all because of the lack of a fine-tuning phase. Based on 19 years of tutorials, the highest accuracy on this dataset is 96.8.

By updating the parameter weights of BERT through fine-tuning, the DistilBERT model can improve the score on the sentence classification task (called the downstream task). In the original tutorial, the accuracy of 91.3 DistilBERT was reached by fine-tuned Distilbert (see huggingface website for the model), and the BERT model with the full number of parameters can achieve a score of 92.7.

Based on the experiment and the research topic "Research on the Application of Deep Learning-based BERT Model in Sentiment Analysis," we can conclude that incorporating the DistilBERT model into sentiment analysis tasks shows promising results. Although the performance achieved in this experiment may not fully demonstrate the potential of BERT due to the lack of fine-tuning, it outperforms traditional word vector models like FastText, Word2Vec, and GloVe. Fine-tuning the DistilBERT model can further enhance its performance, as evidenced by the accuracy improvement achieved in downstream tasks. This suggests that leveraging deep learning-based BERT models holds significant potential for improving sentiment analysis tasks, paving the way for more accurate and efficient sentiment classification in various applications.

## 5. CONCLUSION

Based on the research conducted on the application of Deep Learning-based BERT models in sentiment analysis, several key conclusions can be drawn:

Firstly, the study highlights the significant potential of incorporating BERT models into sentiment analysis tasks. The experiment compared the performance of DistilBERT with traditional word vector models like FastText, Word2Vec, and GloVe. Despite the lack of fine-tuning, DistilBERT showcased promising results, outperforming traditional models. This underscores the effectiveness of leveraging deep learning techniques, particularly BERT models, in enhancing sentiment analysis accuracy and efficiency.

Secondly, while the experiment demonstrated considerable improvements with DistilBERT, it also identified the importance of fine-tuning. By fine-tuning the DistilBERT model, further enhancements in performance were observed, as evidenced by increased accuracy in downstream tasks. This emphasizes the significance of optimizing BERT models for specific sentiment analysis tasks, highlighting the potential for even greater improvements with tailored fine-tuning approaches.

Lastly, the findings suggest that deep learning-based BERT models hold substantial promise for advancing sentiment analysis across various applications. By leveraging the comprehensive understanding of language context offered by BERT's bidirectional Transformer architecture, more accurate and efficient sentiment classification can be achieved. This opens up avenues for better understanding public sentiment, enhancing decision-making processes in areas such as product review analysis, public opinion monitoring, and consumer behavior prediction. Thus, the study underscores the transformative potential of BERT models in revolutionizing sentiment analysis methodologies and driving advancements in natural language processing research.